# Automated Thermal Face recognition based on Minutiae Extraction

Ayan Seal[*,#], Suranjan Ganguly, Debotosh Bhattacharjee, Mita Nasipuri and Dipak Kr. Basu

Department of Computer Science and Engineering,
Jadavpur Uinersity,
188, Raja S.C.Mallick Road, Jadavpur, Kolkata-700032, India
E-mail: ayan.seal@gmail.com, suranjanganguly@gmail.com, debotosh@indiatimes.com, mnasipuri@cse.jdvu.ac.in, dipakkbasu@gmail.com
*Corresponding author
[#]DST INSPIRE FELLOW

**Abstract:** In this paper an efficient approach for human face recognition based on the use of minutiae points in thermal face image is proposed. The thermogram of human face is captured by thermal infra-red camera. Image processing methods are used to pre-process the captured thermogram, from which different physiological features based on blood perfusion data are extracted. Blood perfusion data are related to distribution of blood vessels under the face skin. In the present work, three different methods have been used to get the blood perfusion image, namely bit-plane slicing and medial axis transform, morphological erosion and medial axis transform, sobel edge operators. Distribution of blood vessels is unique for each person and a set of extracted minutiae points from a blood perfusion data of a human face should be unique for that face. Two different methods are discussed for extracting minutiae points from blood perfusion data. For extraction of features entire face image is partitioned into equal size blocks and the total number of minutiae points from each block is computed to construct final feature vector. Therefore, the size of the feature vectors is found to be same as total number of blocks considered. A five layer feed-forward back propagation neural network is used as the classification tool. A number of experiments were conducted to evaluate the performance of the proposed face recognition methodologies with varying block size on the database created at our own laboratory. It has been found that the first method supercedes the other two producing an accuracy of 97.62% with block size 16×16 for bit-plane 4.

**Biographical notes:** Ayan Seal is a DST INSPIRE Fellow of Computer Science and Engineering department at Jadavpur University, India. Currently he is pursuing PhD in Engineering. His main research activities concern the thermal face recognition for biometric security system with special regard to Image Processing and Machine Learning.

Suranjan Ganguly is a Project Fellow, UGC sponsored major research project entitled "Design and Development of Facial Thermogram Technology for Biometric Security System" under the supervision of Dr. Debotosh Bhattacharjee and Prof. Mita Nasipuri of the project, department of Computer Science and Engineering of this university. Currently he is pursuing M.Tech in Computer Technology. His main research activities concern the thermal face recognition for biometric security system with special regard to Image Processing and Machine Learning.

Debotosh Bhattacharjee received the MCSE and Ph. D.(Eng.) degrees from Jadavpur University, India, in 1997 and 2004 respectively. He was associated with different institutes in various capacities until March 2007. After that he joined his Alma Mater, Jadavpur University. His research interests pertain to the applications of computational intelligence techniques like Fuzzy logic, Artificial Neural Network, Genetic Algorithm, Rough Set Theory, Cellular

Automata etc. in Face Recognition, OCR, and Information Security. He is a life member of Indian Society for Technical Education (ISTE, New Delhi), Indian Unit for Pattern Recognition and Artificial Intelligence (IUPRAI), and member of IEEE (USA).

Mita Nasipuri received her B.E.Tel.E., M.E.Tel.E., and Ph.D. (Engg.) degrees from Jadavpur University, in 1979, 1981 and 1990, respectively. Prof. Nasipuri has been a faculty member of J.U since 1987. Her current research interest includes image processing, pattern recognition, and multimedia systems. She is a senior member of the IEEE, U.S.A., Fellow of I.E (India) and W.B.A.S.T, Kolkata, India.

Dipak Kr. Basu received his B.E.Tel.E., M.E.Tel., and Ph.D. (Engg.) degrees from Jadavpur University, in 1964, 1966 and 1969 respectively. Prof. Basu has been a faculty member of J.U from 1968 to January 2008. His current fields of research interest include pattern recognition, image processing, and multimedia systems. He is a senior member of the IEEE, U.S.A., Fellow of I.E. (India) and W.B.A.S.T., Kolkata, India and a former Fellow, Alexander von Humboldt Foundation, Germany.

## 1 Introduction

In present era, people are very much conscious about their protection and safety. So, the authentication technologies, such as use of secret code, magnetic card etc which can be copied, reproduced or stolen are no longer able to meet the present needs. On the other hand, biometric traits are free from such demerits and are getting importance in development of security and authentication systems. The word biometrics is derived from the ancient Greek words "bios" meaning life and "metron" meaning measure (Toth, 2005), (Jain et al., 2007). So, the meaning of biometric is life measurement. Biometrics uses various physical characteristics or personal traits shown in figure 1 to match with the data in the database to identify the candidates. Physical characteristics are suitable for identity verification of human beings and generally obtained from living human body. Commonly used physical biometrics are fingerprints, facial features, hand geometry and eye features (iris and retina) etc. Personal trait is more appropriate in application but with less accuracy because it needs direct physical interaction. The most commonly used personal traits are signature, voice and gait etc. Face recognition has become one of the most attractive research areas among the researchers because of its wide acceptability among various commercial and law enforcement agencies as an important authentication tool. There is a number of reasons to make face biometric to be superior compared to other biometrics. The most important one is that no physical interaction is needed. It is the only biometric that allows you to perform passive detection in different circumstances. The system can recognize the identity of the person without his consent. Since last three decades, there were extensive research activities in face recognition and a good number of such systems are commercially available; still researches are needed to handle the problems due to facial expressions, varying poses, non-uniform illuminations and occlusions. Commonly there are two methods for capturing an image. One is visual imaging and another is thermal imaging. Visual images captured by optical cameras are more common than thermal images captured by infra-red cameras. Advantages of visual images are as follows:

- locating and extraction of features can be done easily and
- optical cameras are not very expensive.

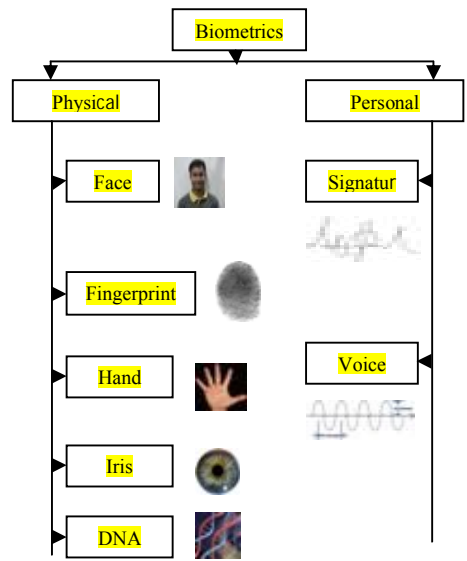

**Figure 1** Biometric uses

Face recognition based on visible images suffers from several problems (Kong et al., 2005) such as:

- illumination conditions,

- viewing directions or poses,
- facial expressions,
- aging, and
- disguises such as facial hair, glasses, or cosmetics.

To overcome this limitation, several solutions have been designed. One solution is using 3D data obtained from 3D vision device. Such systems are less dependent on illumination changes, but again they have some disadvantages: the cost of such system is high and their processing speed is low. Another solution for the above mentioned problem is to use infrared facial images. The word infra is derived from Latin and meaning of this word is below so the infrared means below red. Color of the longest wavelengths of visual spectrum is red. Infrared wave has a longer wavelength (and so a lower frequency) than that of red illumination visible to humans. Objects usually emit infrared radiation across a spectrum of wavelengths, but only a specific region of the spectrum is of interest because sensors are usually designed only to collect radiation within a specific bandwidth. The infrared spectrum is divided into different bandwidths: Near-IR (NIR), Short-wave-IR (SWIR), Medium-wave-IR (MWIR) and Long-wave IR (Thermal IR). The wavelength ranges of different infrared spectrums are shown in Table 1.

**Table 1** Wavelength ranges for different infrared spectrums (M.K. Bhowmik el al, 2011)

| Spectrum | Wavelength range |
|---|---|
| Near-Infrared (NIR) | 0.7-1.0 μm* |
| Short-wave Infrared (SWIR) | 1-3 μm* |
| Mid-wave Infrared MWIR) | 3-5 μm* |
| Thermal Infrared (TIR) | 8-14 μm* |

\* micro meter / micron

Traditional optical cameras use photosensitive silicon that is typically able to determine energy at electromagnetic wavelengths from 0.4μm to just over 1.0μm. Multiple technologies like CCD technology are presently available, with decreasing cost and increasing performance, which are able to measure different regions of the infrared spectrum, as shown in figure 2.

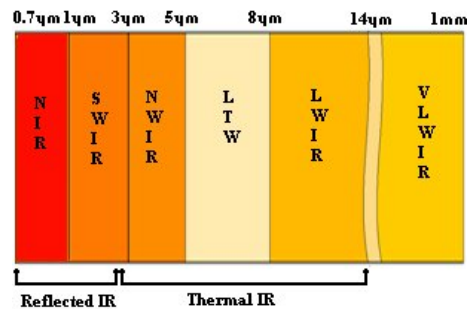

**Figure 2** Different infrared sub-bands of the electromagnetic spectrums

Recently researchers have been using Near-IR imaging cameras for face recognition with better results (Socolinsky et al., 2003), but SWIR and MWIR have not been studied properly till today. Thermal IR camera sensors capture the emitted heat energy from the face and they don't capture the reflected energy. Thermal IR spectrum has some advantages:
- The cost of IR cameras has been considerably reduced with the development of CCD technology (Shiqian et al.).
- Thermal images can be captured under different lighting conditions, even under completely dark environment.
- Thermal IR face images contain basic anatomical information about faces

(Prokoski et al., 2000).
- The tasks of face detection, localization, and segmentation are comparatively easier and more reliable than those in visible band images (Kong et al. 2005).
- Thermal imaging has better accuracy as it uses facial temperature variations caused by vein structure on facial surface as the distinguishing trait.
- As the heat pattern is emitted from the face surface itself without any source of external radiation these systems can capture images despite low illumination or even in the dark. Humans are homoeothermic and hence capable of maintaining constant temperature under different surrounding temperature.

However thermal image of face may get affected due to ambient conditions (Socolinsky et al., 2003), (Chen et al., 2005), (Wu et al., 2007), metabolism (Heo et al., 2005), breathing patterns, and alcohol consumption. Many other factors such as imaging conditions (e.g. distance, glasses etc.), physiological conditions (e.g. toothache, headache etc.) and psychological conditions (e.g. anger, stress etc.) also affect the thermal images. The skin temperature distribution changes from person to person and from time to time (Jones et al., 2002). It is difficult to extract the unique features of a face because complexity remain within the face image (variety of information are stored in the face image). A facial thermal pattern is determined by the vascular structure of each face (Prokoski et al., 2000), which is unique. An infrared camera with good sensitivity can indirectly capture images of superficial blood vessels on the human face (Manohar et al., 2004). However, it has been indicated by Guyton and Hall (Guyton et al., 1996) that the average diameter of blood vessels is around 10-15 μm, which is too small to be detected by current IR cameras because of the limitation in spatial resolution. The skin just above a blood vessel is on an average 0.1 ◦C warmer than the adjacent skin, which is beyond the thermal accuracy of current IR cameras. The convective heat transfer effect from the flow of "hot" arterial blood in superficial vessels creates characteristic thermal imprints, which are at a gradient with the surrounding tissue. Face recognition method does not depend only on the topology of the facial vascular network but also on the fat depositions and skin complexion. The reason is that imagery is formed by the thermal imprints of the vessels and not the vessels directly. Even if the vessel topology was absolutely the same across individuals still, the thermal imprints would differ due to variable absorption from different fat padding (skinny faces versus puffy faces) (Geef et al., 2006). It has been found that there exists an analogy between thermal imprints of human faces and fingerprints of human beings. Thermal imprints of the blood vessels may be treated as the ridges in the fingerprints and fingerprints recognition techniques may be applied on thermal imprints of the human faces for their recognition. The reliability of the fingerprint recognition system has been significantly enhanced through the technique of minutiae extraction from ridges. Most common types of minutiae are: when a ridge either comes to an end, which is called a ridge-termination and when it splits into two ridges, which is called a ridge-bifurcation. Figure 3a illustrates an example of a ridge bifurcation and figure 3b depicts an example of a ridge termination.

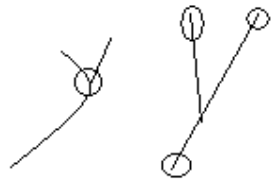

a) Ridge bifurcation b) Ridge termination
**Figure 3** Different types of ridges

The number and locations of the minutiae vary from face to face even for a particular person. When a set of face images is obtained from an individual, the number of minutiae is recorded for each face. The precise locations of the minutiae are also recorded, in the form of numerical coordinates, for each face and the resultant feature vectors thus obtained has been stored in a separate database. A computer can rapidly compare this feature vector with that of anyone else in the world whose face image has been scanned. The paper is organized as follows: Section 2 describes the different steps of proposed approach. Section 3 shows the experimental results and finally, Section 4 concludes the paper.

## 2  The Present System

Thermal Face Recognition System (TFRS) can be subdivided into three main parts. First part is image acquisition, second one is image processing and the third part is classification. The image processing part consists of image binarization from 24-bits color thermal images, extraction of the largest component as a face region, identification of the minutiae points and formation of feature vector. The feature vector is fed into the classification system. Here the classification methodology is based on back propagation feed forward neural network. The block diagram of the present system is given in figure 4. The following subsection describe each of the block in figure 4 in details.

### 2.1  Thermal Face Image Acquisition

A FLIR 7 thermal infra red camera has been used to acquire 24-bits colour thermal face images. The images are saved in JPEG format. A typical thermal face image is shown in figure. This thermal face image depicts interesting thermal information of a facial model. Physiological features of a thermal face image have already been discussed in details in the introduction section.

### 2.2  Binarization

The binarization of 24-bit colour image is divided into two steps. In first step, the colour image is converted into a 8-bit grayscale image using equation 1.

$$I = (0.2989 \times red\_component) + (0.5870 \times green\_component) + (0.1140 \times blue\_component) \ldots \ldots \ldots \ldots \ldots (1)$$

Where 'I' is the grayscale image. The grayscale image corresponding to the thermal image of figure 5 is shown in figure 6.
Grayscale image is then converted into binary image. For this purpose, mean gray value of grayscale image (say gmean) is computed with the help of equation 2.

$$g_{mean} = \frac{\sum_{i=1}^{row} \sum_{j=1}^{column} g(i,j)}{(row \times column)} \ldots (2)$$

If the gray value of any pixel (i, j) (say g(i,j)) is greater than or equal to gmean, then the pixel location in the binary image (i, j) is set with 1 (white) else it is set with 0 (black). The binarization process can be mathematically expressed with the help of equation 3.

$$b(i,j) = \begin{cases} 1 & if\ g(i,j) \geq g_{mean} \\ 0 & otherwise \end{cases} \ldots \ldots \ldots \ldots \ldots (3)$$

In binary image, black pixels mean background and white pixels mean the face region. The binary image corresponding to the grayscale image of figure 6 is shown in figure 7.

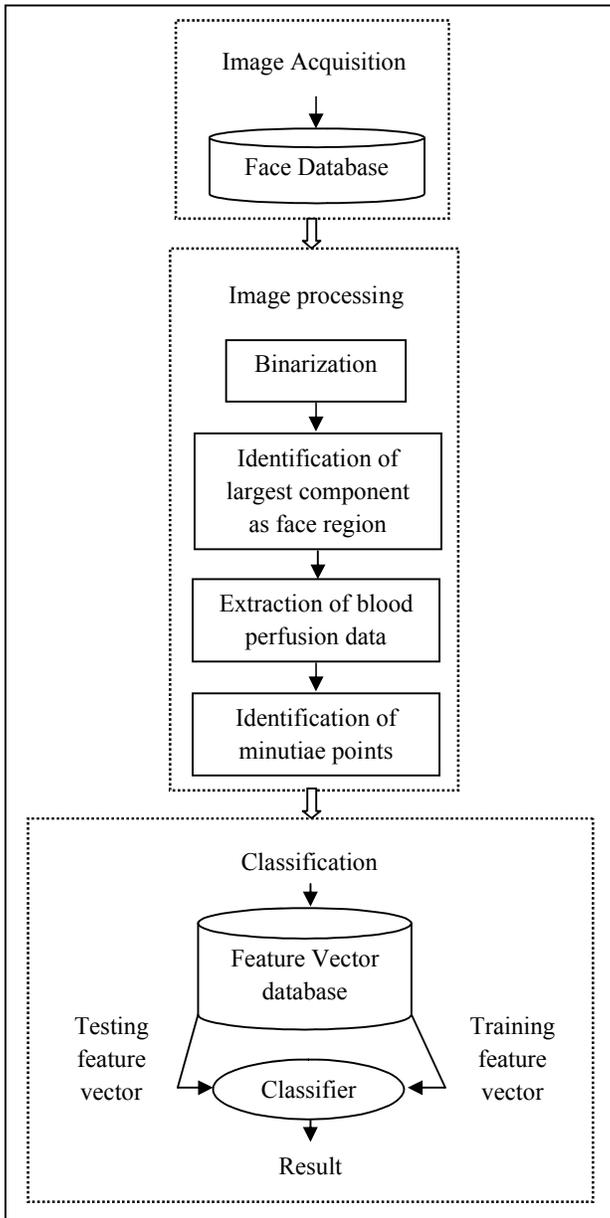

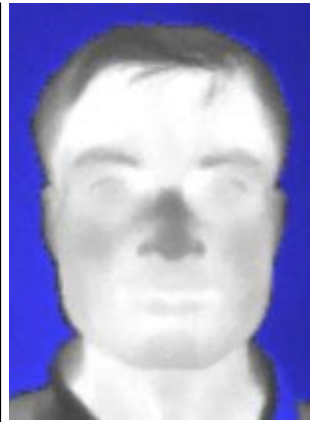

**Figure 5**  A thermal image

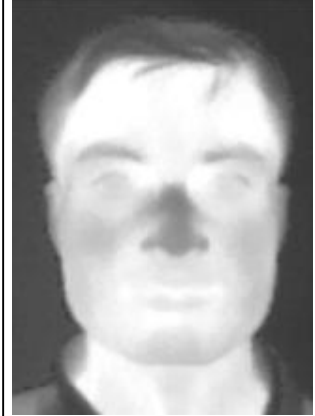

**Figure 6**  Grayscale image

**Figure 4**  Schematic block diagram of the present system

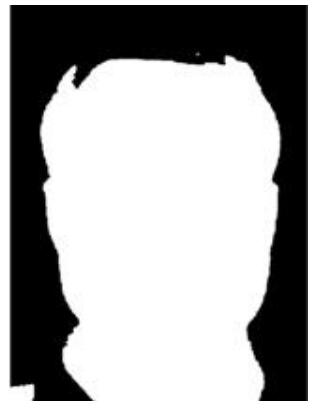

**Figure 7**  Binary image

## 2.3 Identification of Largest Component

The foreground of a binary image may contain more than one object. Say, in figure 7, it has 3 components. The large one represents the face region. The others are at the left hand bottom corner and a small dot on the top. Then largest component has been extracted from binary image using "Connected Component Labeling" algorithm (Bryan et al., 2004). It is based on "4-conneted" neighbours or "8-connected" neighbours method (Gonzalez et al., 2002) and the largest component among them has been identified as face skin region and other small components, which are other parts of the image, have been rejected. There after the image is cropped and unwanted part of the image i.e. the background is eliminated. In binary image, black pixels mean background and white pixels mean the face region. Cropping process starts from top left corner and top right corner of the binary image along the lines and traverses parallel to vertical axis. This process stops when it encounters a white pixel first and then draw a vertical line from two points (one is left side of the face and another is right side of the face), eliminates the left part and right part (i.e. black pixel) of the lines respectively. In the same way it eliminates the upper and lower side of the face region. Figure 8 shows the cropped face region. Then the binary cropped image used as a mask and mapped into grayscale image, to extract the face region from grayscale.

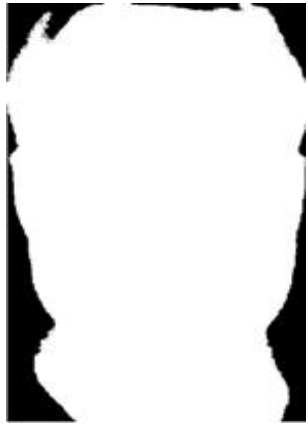 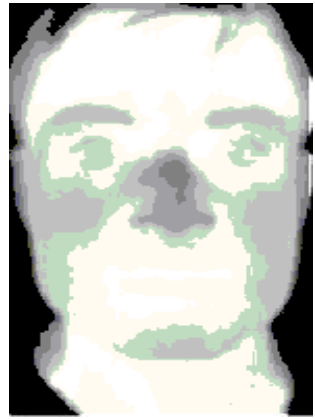

**Figure 8** A cropped face regions

**Figure 9** A cropped grayscale image

## 2.4 Extraction of Blood Perfusion Data

Three different methods are used to extract blood perfusion data from the cropped grayscale image.

- In the first method bit-plane slicing is used to extracted blood perfusion data. This method is used here to extract the details of an image. Bit-plane slicing is a method in which an image is sliced at different planes. Each pixel in a grayscale image is represented by 8 bits that means image is composed of eight 1-bit planes, ranging from bit-plane 0 which is the least significant bit (LSB) up to bit-plane 7 which is the most significant bit (MSB) as shown in figure 10.

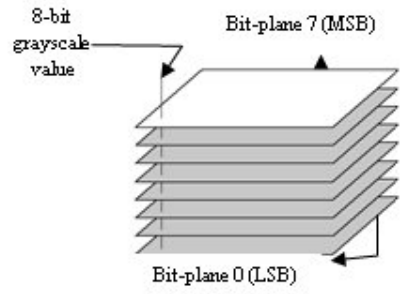

**Figure 10**   Bit-plane representation of an 8-bit grayscale image

Figure 11(a-h) shows the different bit-planes for the grayscale image shown in figure 9.

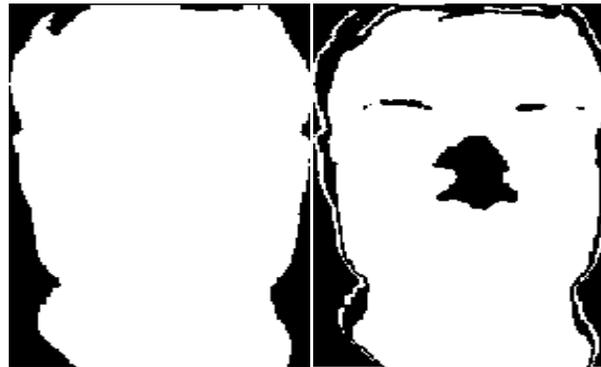

a)   Bit plane 0           b)   Bit plane 1

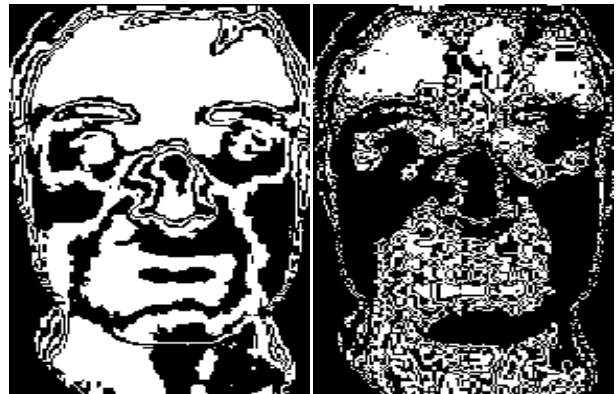

c)   Bit plane 2           d)   Bit plane 3

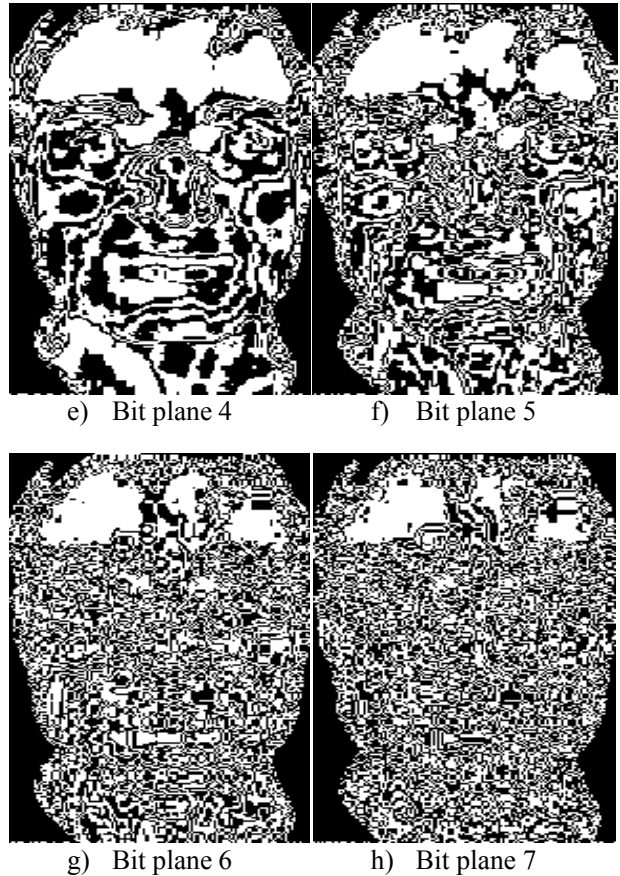

| e) Bit plane 4 | f) Bit plane 5 |
| --- | --- |
| g) Bit plane 6 | h) Bit plane 7 |

**Figure 11** Different bit plane images of figure 9

The advantage of this method is to get the relative importance played by each bit in a pixel. It highlights the contribution made to total image appearance by specific bits. In this method, only 4 higher order bit-planes contain the majority of the visually significant data (Gonzalez et al., 2002). The lower level bit planes do not give much detail because they are made up of lower contrast.

So, bit-plane slicing technique is used here to extract the thermal physiological facial features from a gray scale image and construct the contour maps of regions having constant or equal temperatures. Medial axis transform (Gonzalez et al., 2002) is applied to extract lines similar to isothermal lines in weather maps linking all points of equal or constant temperature in order to get blood perfusion image. Figure 12 shows blood perfusion image for the image shown in Fig 11e.

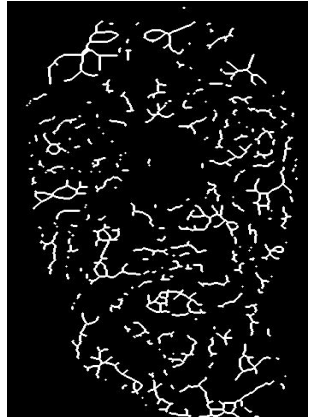

**Figure 12** Blood perfusion image of figure 11e obtained by applying medial axis transform on bit-plane 4 data

- In the second method, morphological gray level erosion (Gonzalez et al., 2002) is used to extract the thermal physiological face features and construct the region having constant or equal temperatures. Medial axis transform is applied as before to extract the blood perfusion data of figure 9, which is shown in figure 13.

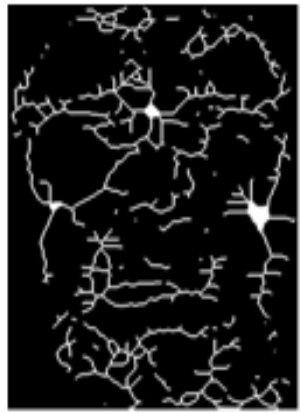

**Figure 13** Blood perfusion image of figure 9 obtained by applying morphological gray level erosion and medical axis transform

- And in the third method "Sobel" operator is used to extract the blood perfusion data. Extraction of the blood perfusion data refers to the process of identifying the discontinuities, in an image. The discontinuities considered here are abrupt changes in pixel intensity. Here "Sobel" edge detector operator (Gonzalez et al., 2002) is used to convolve the image for identifying discontinuities in pixel intensity. The operator consists of a pair of 3×3 convolution masks as shown in Fig. 14. One mask is simply the other flipped by 90º

| -1 | 0 | +1 |
|----|---|----|
| -2 | 0 | +2 |
| -1 | 0 | +1 |

Px

| +1 | +2 | +1 |
|----|----|----|
| 0  | 0  | 0  |
| -1 | -2 | -1 |

Py

**Figure 14** "Sobel" edge detector masks

These masks are designed to respond maximal discontinues in intensity values running vertically and horizontally relative to the pixel grid, one mask for each of the two perpendicular orientations. The masks are applied separately to the input image, to produce separate measurement of the gradient component in each orientation (denoted at Px and Py). They are combined together to find the absolute magnitude of the gradient at each point and the orientation of that gradient. The gradient magnitude is given by equation 4 or 5:

$$|P| = \sqrt{Px^2 + Py^2} \quad \ldots \ldots \ldots (4)$$
$$|P| = |Px| + |Py| \quad \ldots \ldots \ldots (5)$$
$$\theta = \arctan\left(\frac{Py}{Px}\right) \ldots \ldots \ldots (6)$$

The orientation of the gradient is given by equation 6. "Sobel" edge detector masks are applied to extract lines similar to isothermal lines in weather maps linking all points of equal or constant temperature in order to get blood perfusion image. Figure 15 shows blood perfusion data for the image shown in figure 9.

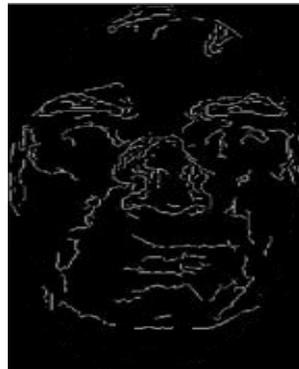

**Figure 15** Blood perfusion image of figure 9 obtained by applying Sobel operator

## 2.5 Identification of Minutiae Points

The idea of minutiae extraction from finger print recognition have been taken from (Jain et al., 2011), (Galton, 1892), (Maltani et al., 2009) and applied it in the present work. There exists an analogy between thermal imprints of human faces and fingerprints of human beings. Here fingerprint's ridges are like blood perfusion data of a face. The uniqueness of a face's blood perfusion data can be determined by the pattern of ridges as well as the minutiae points. Minutiae points are local ridge characteristics that occur either at a ridge bifurcation or at a ridge termination. Before finding the minutiae points, all blood perfusion images could be normalized into same sized images. In our case it is 320×224. For finding the minutiae points the blood perfusion image is scanned from left to right and top to bottom using a 3x3 window. For each position of the window, the numbers of '1's within the window are checked. If the window contains a '1' at its central cell position which has a single '1' as its neighbours as shown, then 16a the central cell position represents a termination point. If the central cell contains '1' and has three neighboring cells with '1', then it is a bifurcation which is shown in figure 16b. If the central cell has value '1' and has two neighbors containing '1's then it is a normal point, shown in figure 16c. Due to noises in the blood perfusion image, the minutiae extraction algorithm produced a large number of spurious minutiae points. Therefore, differentiating spurious minutiae from real minutiae in the post-processing stage is crucial for accurate face recognition. The more spurious minutiae are eliminated; the better will be the classification performance. In addition, classification time will be significantly reduced because reduction of feature points Figure 18 illustrates minutiae points of figure 13, which are basically the bifurcation and termination points represented by green and red colors respectively.

| 0 | 0 | 1 |   | 1 | 1 | 0 |   | 0 | 1 | 0 |
|---|---|---|---|---|---|---|---|---|---|---|
| 0 | 1 | 0 |   | 1 | 1 | 0 |   | 0 | 1 | 1 |
| 0 | 0 | 0 |   | 0 | 0 | 0 |   | 0 | 0 | 0 |

a) Termination point    b) Bifurcation point    c) Normal point

**Figure 16** Binary number indicating the minutiae point

Here, another technique is also used to extract the minutiae points. This technique is based on crossing number (Maltoni el al., 2003). In order to get the crossing number, a 3×3 window is moved from the pixel in row one and column one, and then continues to move to the adjacent column on the right. When the scanning process reaches the last column of the image, it moves one row down and also resets to the first column and continues until the pixel in last row and last column of the whole thinned image. Crossing number finding process starts by looking at the eight sets of adjacent pixels surrounded by a black centered pixel denoted by '0' in bold letter in the Figure 17 and surrounded 8-pixels are denoted by 1 to 8 sequential numbers.

| 2 | 3 | 4 |
|---|---|---|
| 1 | **0** | 5 |
| 8 | 7 | 6 |

**Figure 17** Eight sets of adjacent pixels used in computing crossing number

The difference between two adjacent pixels of the same color is equal to zero and the difference between two adjacent pixels of the different color is equal to one. This difference is individually computed for the eight sets of adjacent pixels and adds them together. The resultant sum is divided by two and stores it at the center of the 3×3 window. When the crossing number is one, then the center pixel is treated as a ridge termination point. In the same way, when the crossing number is greater than or equal to three, then the center pixel is denoted by ridge bifurcation point.

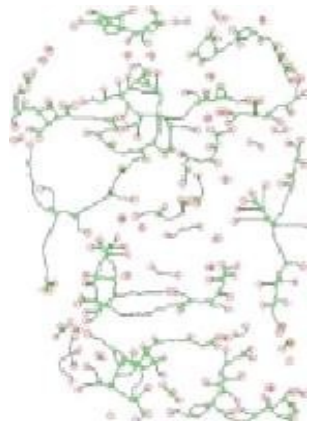

**Figure 18** Minutiae points

## 2.6 Creation of Feature Vector

Since the number of minutiae points in the blood perfusion data varies from person to person, feature vectors formed with the locations of minutiae points will have different lengths for different persons. To overcome this problem, the whole image is divided into number of fixed sized blocks. The size of the each block may be 8×8, 16×16, and 32×32. The number of minutiae points in each block have been counted and stored in a vector. For each face image, one fixed length feature vector is thus formed. The total number of such vectors is equal to total number of faces.

### 2.7 Classification of Blood Perfusion Images

These feature vectors are divided into two sets one for training purpose and another for testing purpose. ANN classifier has been used to classify each of these vectors (Turk et al., 1991), (Lin et al., 1996), (Haddadnia et al., 2003). Here a five layer feed-forward back propagation neural network has been used for classification purpose since is one of the most popular neural networks to the scientific and engineering community for modeling and processing of many quantitative phenomena. Momentum allows the network to respond to local gradient and to recent trends in the error surface. The momentum is used to back propagation learning algorithm for making weight changes equal to the sum of a fraction of the last weight change and the new change. The magnitude of the effect that the last weight change is allowed is known as momentum constant (mc). The momentum constant may be any number between 0 and 1. The momentum constant zero means, a weight changes according to the gradient and the momentum constant one means, the new weight change is set to equal the last weight change, and the gradient is not considered here. The gradient is computed by summing the gradients calculated at each training example, and the weights and biases are only updated after all training examples have been presented. Tan-sigmoid transfer functions are used to calculate a layer's output from its net, the first input, and the next three hidden layers and the outer most layer gradient descent with momentum training function is used to update weight and bias values. First hidden layer contains 100 neurons, second hidden layer contains 50 neurons, third hidden layer holds 10 neurons, input layer contains 238 neurons for training of 238 images and the last layer contains 7 neurons because 7 different persons are there in our experiment.

### 3  Experiment and Result

The proposed algorithm is tested on thermal face images captured using a FLIR 7 camera at our own laboratory. Some preprocessing has been done for the image database used in this paper. All the training and testing images are in color with 24-bits. Face images of a person was captured in normal temperature conditions, and each person sitting on a chair at a distance of about 2 feet in front of the thermal IR camera. Since objects within a limited range of distances from the camera will be reproduced clearly. We have calculated the distance that is 2 feet by adjusting the camera movement to the subjects which helped us to capture thermal face images clearly. Till now we have captured images of 17 persons. Each person has 34 different templates: emotion type without changing head orientation, different views about y axis, different views about x axis, and different views about z axis. When a single test is performed, the person may belong to a particular class or the person may not belong to a particular class. The test result may be positive, indicating the belongingness to a particular class, or the test result may be negative, indicating the image does not belong to a class. The table 2 shows test results in the rows and true status of the person being plotted in the columns.

Table 2 Result vs. true status of nature

| True status of nature (B) / Result (A) | Does belong to particular class(+) | Does not belong to particular class (-) |
|---|---|---|
| (+) | Tp | Fp |
| (-) | Fn | Tn |

The biometric security systems generally uses two principal criteria to access the accuracy. These are: "sensitivity" and "specificity". Sensitivity as the probability that the test says a person belongs to a particular class and actually that one belongs of that particular class. That is

$$P(A+/B+) = T_p/(T_p + F_n) \quad \ldots \ldots \ldots (7)$$

Specificity as the probability that the test says a person does not belong to a particular class when in fact they belong to that class. This is

$$P(A-/B-) = T_n/(F_p + T_n) \quad \ldots \ldots \ldots (8)$$

Generally, a test should have high sensitivity and high specificity. Sometimes a test gives high sensitivity and low specificity then we go for false acceptance rate (FAR)- the FAR is the measurement of the probability that the test reports a positive result for a person who is not in a particular class. The false acceptance rate is given by-

$$P(B-/A+) = F_p/(F_p + T_n) = 1 - \text{specifity} \quad \ldots \ldots \ldots (9)$$

False rejection rate (FRR)- the FRR is the measurement of the probability that the test reports a negative result for a person who actually belong to a particular class. The false rejection rate is given by

$$P(B+/A-) = F_n/(T_p + F_n) = 1 - \text{sensitivity} \quad \ldots \ldots \ldots (10)$$

The accuracy of the system is represented by

$$\text{Accuracy} = (T_p + T_n)/(T_p + T_n + F_p + F_n) \quad \ldots \ldots (11)$$

### 3.1 Experiment 1:

In the first method, bit-plane slicing and medial axis have been used to extract blood perfusion data. Two different methods are discussed in section 2.5 for extraction of minutiae points from blood perfusion data. Then the minutiae point's image is divided into 8×8, 16×16 and 32×32 blocks and the total number of minutiae points in each block is counted and stored in a one dimensional vector. So, the total number of features are 1120, 280 and 70 for 8×8, 16×16 and 32×32.the block sizes respectively. The obtained results are shown in Table 3 for different block sizes. Table 3 shows the performance rate and computation time in seconds for verifying whether a test face is in the correct class or not.

**Table 3** variation of performance rate with block sizes for different bit-plane of blood perfusion image

|  | Size of the block | Performance rate (%) | | computational time in seconds | |
| --- | --- | --- | --- | --- | --- |
|  |  | Using 1st set of minutiae points | Using 2nd set of minutiae points | Using 1st set of minutiae points | Using 2nd set of minutiae points |
| **Bit-plane 0** | 8×8 | 85.71 | 76.19 | 0.031247 | 0.023528 |
|  | 16×16 | 95.24 | 78.57 | 0.018115 | 0.111777 |
|  | 32×32 | 85.71 | 80.95 | 0.028890 | 0.015917 |
| **Bit-plane 1** | 8×8 | 85.71 | 85.71 | 0.015789 | 0.041948 |
|  | 16×16 | 95.24 | 78.57 | 0.025623 | 0.018880 |
|  | 32×32 | 95.24 | 78.57 | 0.020862 | 0.015670 |
| **Bit-plane 2** | 8×8 | 85.71 | 71.42 | 0.047827 | 0.030623 |
|  | 16×16 | 85.71 | 83.33 | 0.016452 | 0.019862 |
|  | 32×32 | 80.95 | 76.19 | 0.108189 | 2.154909 |
| **Bit-plane 3** | 8×8 | 85.71 | 80.95 | 0.015756 | 0.040217 |
|  | 16×16 | 85.71 | 76.19 | 0.067336 | 0.017803 |
|  | 32×32 | 90.48 | 83.33 | 0.021935 | 0.016102 |
| **Bit-plane 4** | 8×8 | 90.48 | 76.19 | 0.032842 | 0.038091 |
|  | 16×16 | 95.24 | 69.04 | 0.035008 | 0.017884 |
|  | 32×32 | 95.24 | 80.95 | 0.017265 | 0.037117 |
| **Bit-plane 5** | 8×8 | 95.24 | 83.33 | 0.015235 | 0.049629 |
|  | 16×16 | 80.95 | 90.48 | 0.012563 | 0.018402 |
|  | 32×32 | 90.48 | 80.95 | 0.041897 | 0.032391 |
| **Bit-plane 6** | 8×8 | 80.95 | 78.57 | 0.041211 | 0.098166 |
|  | 16×16 | 80.95 | 80.95 | 0.051114 | 0.017762 |
|  | 32×32 | 76.19 | 83.33 | 2.006090 | 0.021837 |
| **Bit-plane 7** | 8×8 | 85.71 | 78.57 | 0.208960 | 0.044336 |
|  | 16×16 | 85.71 | 85.71 | 0.225659 | 0.051452 |
|  | 32×32 | 80.95 | 85.71 | 0.151052 | 0.015590 |

So, the maximum success of 95.24% recognition has been achieved with block size 16×16 and 32×32 with bit-plane 4, given in table 3 and accuracy rate of 97.62% has been achieved with block size 16×16 for bit-plane 4, given in table 4. The experimental results illustrate that the proposed algorithm is much better for bit-plane 4 than other bit-planes.

**Table 4** variation of FAR, FRR and accuracy with block sizes for bit-planes of blood perfusion images

|  | Size of the block | FAR (%) | | FRR (%) | | Accuracy (%) | |
| --- | --- | --- | --- | --- | --- | --- | --- |
|  |  | Using 1st set of minutiae points | Using 2nd set of minutiae points | Using 1st set of minutiae points | Using 2nd set of minutiae points | Using 1st set of minutiae points | Using 2nd set of minutiae points |
| **Bit-plane 0** | 8×8 | 14.29 | 14.29 | 14.29 | 23.81 | 85.71 | 80.95 |
|  | 16×16 | 19.05 | 14.29 | 04.76 | 23.81 | 88.10 | 80.95 |
|  | 32×32 | 19.05 | 19.05 | 14.29 | 19.05 | 83.33 | 80.95 |
| **Bit-plane 1** | 8×8 | 14.29 | 09.52 | 14.29 | 14.29 | 85.71 | 88.10 |
|  | 16×16 | 23.81 | 14.29 | 04.76 | 23.81 | 85.71 | 80.95 |
|  | 32×32 | 19.05 | 14.29 | 04.76 | 23.81 | 88.10 | 80.95 |
| **Bit-plane 2** | 8×8 | 09.52 | 14.29 | 14.29 | 23.81 | 88.10 | 80.95 |
|  | 16×16 | 23.81 | 19.05 | 14.29 | 19.05 | 80.95 | 80.95 |
|  | 32×32 | 19.05 | 19.05 | 19.05 | 23.81 | 80.95 | 78.57 |
| **Bit-plane 3** | 8×8 | 09.52 | 14.29 | 14.29 | 14.29 | 88.10 | 85.71 |
|  | 16×16 | 14.29 | 19.05 | 14.29 | 23.81 | 85.71 | 78.57 |
|  | 32×32 | 14.29 | 19.05 | 14.29 | 14.29 | 85.71 | 80.95 |
| **Bit-plane 4** | 8×8 | 14.29 | 19.05 | 04.76 | 23.81 | 90.48 | 78.57 |
|  | 16×16 | 00.00 | 14.29 | 09.52 | 14.29 | 97.62 | 80.95 |
|  | 32×32 | 09.52 | 14.29 | 04.76 | 19.05 | 92.86 | 85.71 |
| **Bit-plane 5** | 8×8 | 14.29 | 19.05 | 04.76 | 14.29 | 90.48 | 85.71 |
|  | 16×16 | 19.05 | 09.52 | 19.05 | 19.05 | 80.95 | 85.71 |
|  | 32×32 | 09.52 | 19.05 | 09.52 | 14.29 | 90.48 | 80.95 |
| **Bit-plane 6** | 8×8 | 09.52 | 14.29 | 19.05 | 23.81 | 85.71 | 80.95 |
|  | 16×16 | 14.29 | 14.29 | 19.05 | 19.05 | 83.33 | 83.33 |
|  | 32×32 | 14.29 | 19.05 | 23.81 | 14.29 | 80.95 | 85.71 |
| **Bit-plane 7** | 8×8 | 14.29 | 14.29 | 14.29 | 23.81 | 85.71 | 80.95 |
|  | 16×16 | 19.05 | 19.05 | 14.29 | 14.29 | 83.33 | 83.33 |

| | | | | | | |
|---|---|---|---|---|---|---|
| | 32×32 | 19.05 | 19.05 | 19.05 | 14.29 | 80.95 | 83.33 |

## 3.2 Experiment 2:

In the second method, morphological erosion and medial axis have been used to extract blood perfusion data. Here three different masks are used to extract minutiae points (e.g. termination point, bifurcation point and normal point), which have been discussed details in section 2.5. Then the minutiae point's image is divided into 8×8, 16×16 and 32×32 blocks and the total number of minutiae points in each block is counted and stored in a one dimensional vector. So, the total number of features are 1120, 280 and 70 for 8×8, 16×16 and 32×32.the block sizes respectively. Table 5 shows the verification accuracy of the proposed algorithm. So, the maximum recognition performance of 90.48% has been achieved with block size 8×8, and the corresponding accuracy rate 90.48% has been achieved with block size 8×8 also.

**Table 5** variation of performance rate FAR, FRR and accuracy with different block sizes for the second method.

| | No of Block | Performance rate | FAR (%) | FRR (%) | Accuracy (%) | computational time in seconds |
|---|---|---|---|---|---|---|
| **Using 1st set of minutiae points** | 8×8 | 90.48 | 14.29 | 04.76 | 90.48 | 0.026918 |
| | 16×16 | 76.19 | 14.29 | 23.81 | 80.95 | 0.210892 |
| | 32×32 | 85.71 | 14.29 | 14.29 | 85.71 | 0.017936 |
| **Using 2nd set of minutiae points** | 8×8 | 95.24 | 00.00 | 14.29 | 92.86 | 0.136307 |
| | 16×16 | 83.33 | 19.05 | 14.29 | 83.33 | 0.021422 |
| | 32×32 | 78.57 | 14.29 | 23.81 | 80.95 | 0.017503 |

## 3.3 Experiment 3:

In the third method, "Sobel" operator has been used to extract blood perfusion data. Here three different masks are used to extract minutiae points (e.g. termination point, bifurcation point and normal point), which have been discussed details in section 2.5. Then the minutiae point's image is divided into 8×8, 16×16 and 32×32 blocks and the total number of minutiae points in each block is counted and stored in a one dimensional vector. So, the total number of features are 1120, 280 and 70 for 8×8, 16×16 and 32×32.the block sizes respectively. Table 6 shows the verification accuracy of the proposed algorithm. So, the maximum success of 90.48% recognition has been achieved with block size 8×8, and accuracy rate of 90.48% has been achieved with block size 8×8 also.

**Table 6** variation of performance rate FAR, FRR and accuracy with different block sizes for the third method.

| | No of Block | Performance rate | FAR (%) | FRR (%) | Accuracy (%) | computational time in seconds |
|---|---|---|---|---|---|---|
| **Using 1st set of minutiae points** | 8×8 | 90.48 | 14.29 | 04.76 | 90.48 | 0.010010 |
| | 16×16 | 85.71 | 14.29 | 14.29 | 85.71 | 0.323892 |
| | 32×32 | 85.71 | 14.29 | 14.29 | 85.71 | 0.057366 |
| **Using 2nd set of minutiae points** | 8×8 | 85.71 | 14.29 | 14.29 | 85.71 | 0.145690 |
| | 16×16 | 85.71 | 14.29 | 14.29 | 85.71 | 0.126584 |
| | 32×32 | 85.71 | 14.29 | 14.29 | 85.71 | 0.658702 |

## 4   Conclusion

Minutiae based thermal face recognition using blood perfusion data has been proposed here. Three different methods are used to extract blood perfusion data. Then entire face image is divided into equal number of blocks and the total number of minutiae points from each block is considered as one feature. Features from all the blocks are combined to create the final feature vector. Classification of these feature vectors has been done using a multilayer perceptron. Final recognition rate has been enhanced by varying size of the blocks. So, the maximum success of 95.24% recognition has been achieved with block size 16×16 and 32×32 with bit-plane 4 using first method that is bit-plane slicing method. One of the major advantages of this approach is the ease of implementation. Furthermore, no knowledge of geometry or specific feature of the face is required. However, this system is applicable to front views and constant background only. It may fail in unconstraint environments like natural scenes.

**Acknowledgements**

Authors are thankful to a major project entitled "Design and Development of Facial Thermogram Technology for Biometric Security System," funded by University Grants Commission (UGC),India and "DST-PURSE Programme" at Department of Computer Science and Engineering, Jadavpur University, India for providing necessary infrastructure to conduct experiments relating to this work. Ayan Seal is grateful to Department of Science & Technology (DST), India for providing him Junior Research Fellowship-Professional (JRF-Professional) under DST-INSPIRE Fellowship programme [No: IF110591].

**Figure 1** Biometric uses

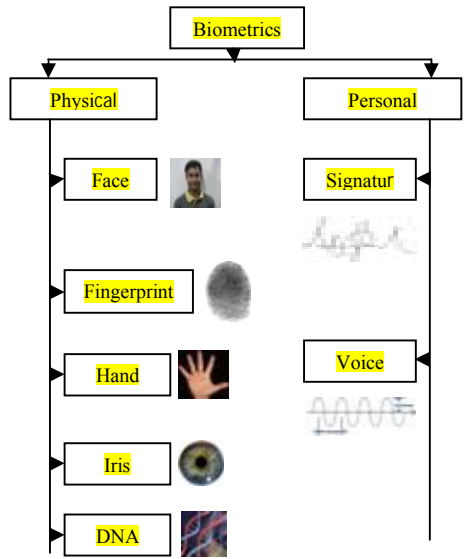

**Table 1** Wavelength ranges for different infrared spectrums (M.K. Bhowmik el al, 2011)

| Spectrum | Wavelength range |
|---|---|
| Near-Infrared (NIR) | 0.7-1.0 μm* |
| Short-wave Infrared (SWIR) | 1-3 μm* |
| Mid-wave Infrared MWIR) | 3-5 μm* |
| Thermal Infrared (TIR) | 8-14 μm* |

\* micro meter / micron

**Figure 2** Different infrared sub-bands of the electromagnetic spectrums

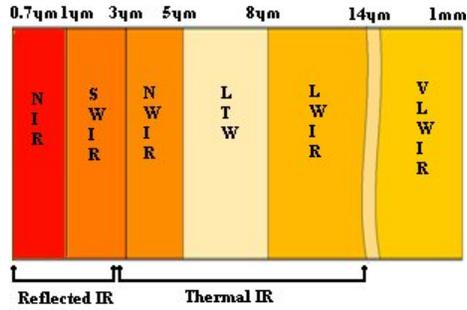

**Figure 3** Different types of ridges

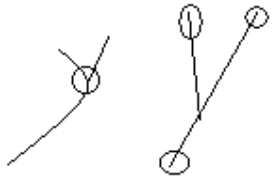

a) Ridge bifurcation   b) Ridge termination

**Figure 4** Schematic block diagram of the present system

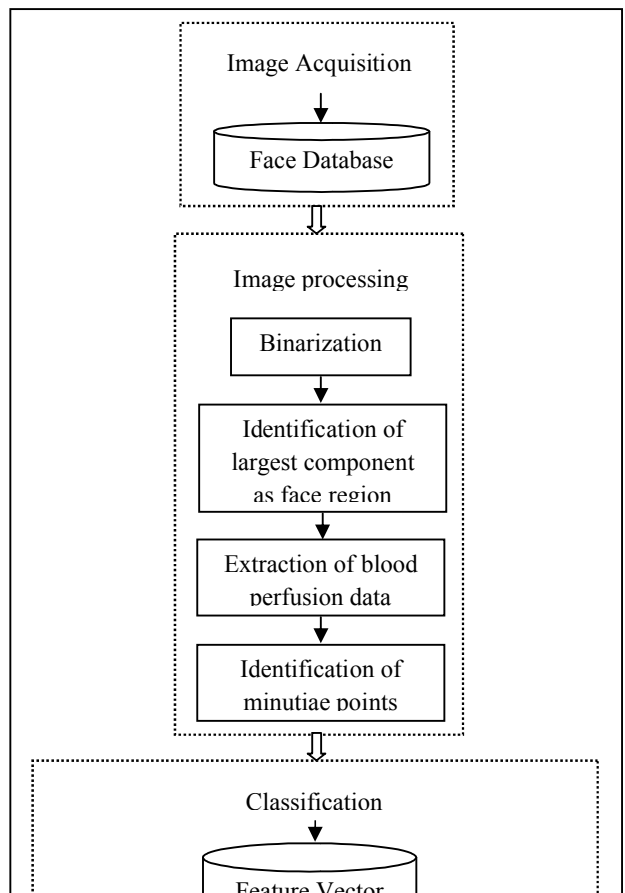

**Figure 5** A thermal image

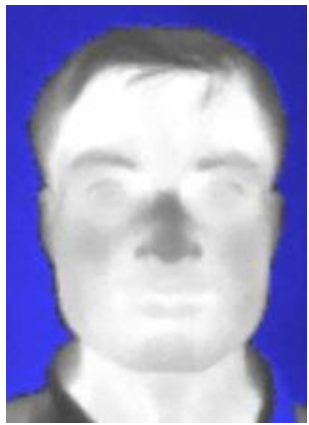

**Figure 6** Grayscale image

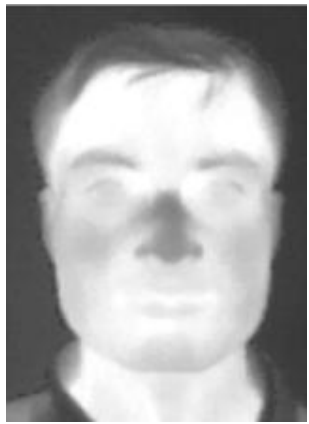

**Figure 7** Binary image

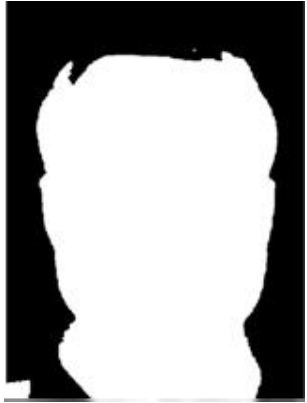

**Figure 8** A cropped face regions

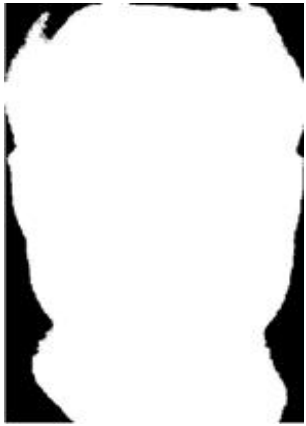

**Figure 9** A cropped grayscale image

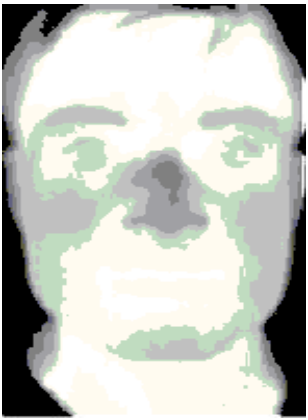

**Figure 10**  Bit-plane representation of an 8-bit grayscale image

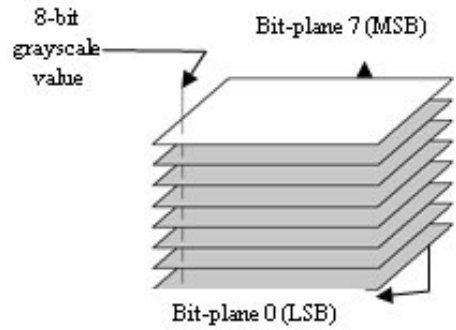

**Figure 11** Different bit plane images of figure 9

Figure 11(a-h) shows the different bit-planes for the grayscale image shown in figure 9.

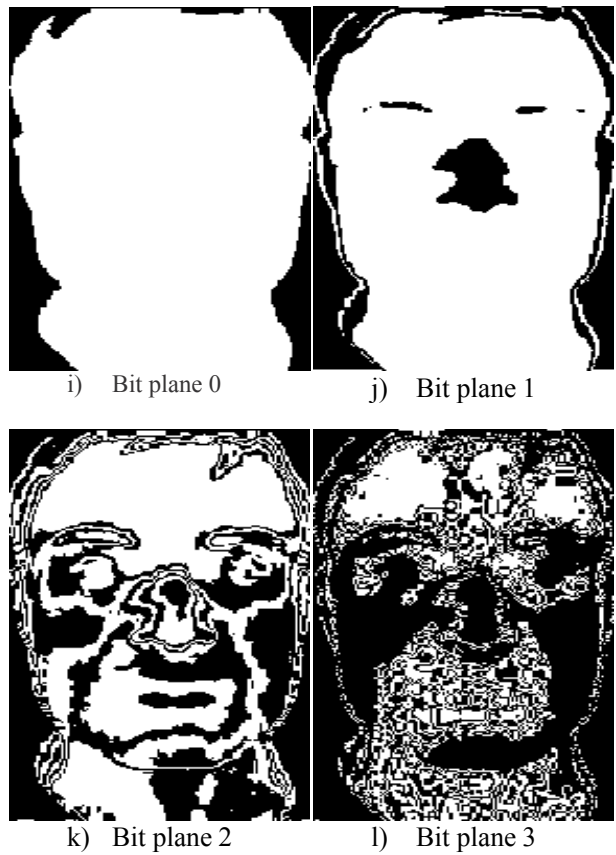

i)  Bit plane 0        j)  Bit plane 1

k)  Bit plane 2        l)  Bit plane 3

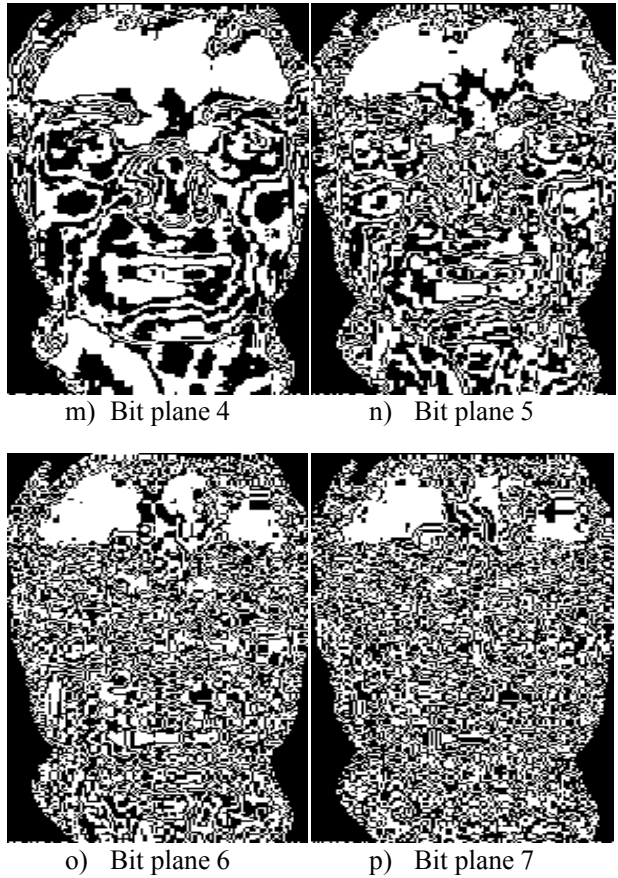

m)  Bit plane 4      n)  Bit plane 5

o)  Bit plane 6      p)  Bit plane 7

**Figure 12** Blood perfusion image of figure 11e obtained by applying medial axis transform on bit-plane 4 data

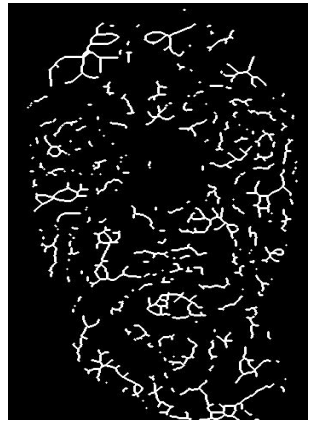

**Figure 13** Blood perfusion image of figure 9 obtained by applying morphological gray level erosion and medical axis transform

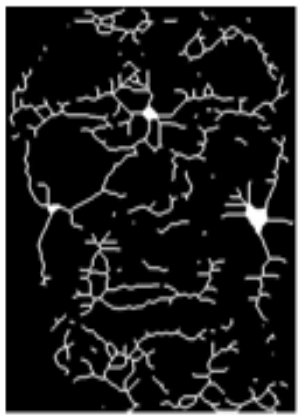

Figure 14 "Sobel" edge detector masks

| -1 | 0 | +1 |
|----|---|----|
| -2 | 0 | +2 |
| -1 | 0 | +1 |

Px

| +1 | +2 | +1 |
|----|----|----|
| 0  | 0  | 0  |
| -1 | -2 | -1 |

Py

Figure 15 Blood perfusion image of figure 9 obtained by applying Sobel operator

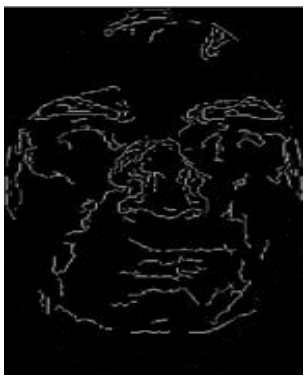

Figure 16 Binary number indicating the minutiae point

| 0 | 0 | 1 |
|---|---|---|
| 0 | 1 | 0 |
| 0 | 0 | 0 |

a) Termination point

| 1 | 1 | 0 |
|---|---|---|
| 1 | 1 | 0 |
| 0 | 0 | 0 |

b) Bifurcation point

| 0 | 1 | 0 |
|---|---|---|
| 0 | 1 | 1 |
| 0 | 0 | 0 |

c) Normal point

**Figure 17** Eight sets of adjacent pixels used in computing crossing number

| 2 | 3 | 4 |
|---|---|---|
| 1 | **0** | 5 |
| 8 | 7 | 6 |

**Figure 18** Minutiae points

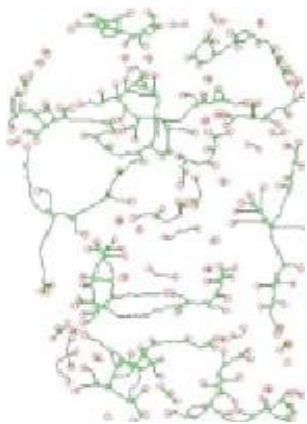

**Table 2** Result vs. true status of nature

| True status of nature (B) / Result (A) | Does belong to particular class(+) | Does not belong to particular class (-) |
|---|---|---|
|  |  |  |

|     | (+) | Tp | Fp |
| --- | --- | --- | --- |
|     | (-) | Fn | Tn |

**Table 3** variation of performance rate with block sizes for different bit-plane of blood perfusion image

|  | Size of the block | Performance rate (%) | | computational time in seconds | |
| --- | --- | --- | --- | --- | --- |
|  |  | Using 1st set of minutiae points | Using 2nd set of minutiae points | Using 1st set of minutiae points | Using 2nd set of minutiae points |
| **Bit-plane 0** | 8×8 | 85.71 | 76.19 | 0.031247 | 0.023528 |
|  | 16×16 | 95.24 | 78.57 | 0.018115 | 0.111777 |
|  | 32×32 | 85.71 | 80.95 | 0.028890 | 0.015917 |
| **Bit-plane 1** | 8×8 | 85.71 | 85.71 | 0.015789 | 0.041948 |
|  | 16×16 | 95.24 | 78.57 | 0.025623 | 0.018880 |
|  | 32×32 | 95.24 | 78.57 | 0.020862 | 0.015670 |
| **Bit-plane 2** | 8×8 | 85.71 | 71.42 | 0.047827 | 0.030623 |
|  | 16×16 | 85.71 | 83.33 | 0.016452 | 0.019862 |
|  | 32×32 | 80.95 | 76.19 | 0.108189 | 2.154909 |
| **Bit-plane 3** | 8×8 | 85.71 | 80.95 | 0.015756 | 0.040217 |
|  | 16×16 | 85.71 | 76.19 | 0.067336 | 0.017803 |
|  | 32×32 | 90.48 | 83.33 | 0.021935 | 0.016102 |
| **Bit-plane 4** | 8×8 | 90.48 | 76.19 | 0.032842 | 0.038091 |
|  | 16×16 | 95.24 | 69.04 | 0.035008 | 0.017884 |
|  | 32×32 | 95.24 | 80.95 | 0.017265 | 0.037117 |
| **Bit-plane 5** | 8×8 | 95.24 | 83.33 | 0.015235 | 0.049629 |
|  | 16×16 | 80.95 | 90.48 | 0.012563 | 0.018402 |
|  | 32×32 | 90.48 | 80.95 | 0.041897 | 0.032391 |
| **Bit-plane 6** | 8×8 | 80.95 | 78.57 | 0.041211 | 0.098166 |
|  | 16×16 | 80.95 | 80.95 | 0.051114 | 0.017762 |
|  | 32×32 | 76.19 | 83.33 | 2.006090 | 0.021837 |
| **Bit-plane 7** | 8×8 | 85.71 | 78.57 | 0.208960 | 0.044336 |
|  | 16×16 | 85.71 | 85.71 | 0.225659 | 0.051452 |
|  | 32×32 | 80.95 | 85.71 | 0.151052 | 0.015590 |

**Table 4** variation of FAR, FRR and accuracy with block sizes for bit-planes of blood perfusion images

|  | Size of the block | FAR (%) | | FRR (%) | | Accuracy (%) | |
| --- | --- | --- | --- | --- | --- | --- | --- |
|  |  | Using 1st set of minutiae points | Using 2nd set of minutiae points | Using 1st set of minutiae points | Using 2nd set of minutiae points | Using 1st set of minutiae points | Using 2nd set of minutiae points |
| **Bit-plane 0** | 8×8 | 14.29 | 14.29 | 14.29 | 23.81 | 85.71 | 80.95 |
|  | 16×16 | 19.05 | 14.29 | 04.76 | 23.81 | 88.10 | 80.95 |
|  | 32×32 | 19.05 | 19.05 | 14.29 | 19.05 | 83.33 | 80.95 |
| **Bit-plane 1** | 8×8 | 14.29 | 09.52 | 14.29 | 14.29 | 85.71 | 88.10 |
|  | 16×16 | 23.81 | 14.29 | 04.76 | 23.81 | 85.71 | 80.95 |
|  | 32×32 | 19.05 | 14.29 | 04.76 | 23.81 | 88.10 | 80.95 |
| **Bit-plane 2** | 8×8 | 09.52 | 14.29 | 14.29 | 23.81 | 88.10 | 80.95 |
|  | 16×16 | 23.81 | 19.05 | 14.29 | 19.05 | 80.95 | 80.95 |
|  | 32×32 | 19.05 | 19.05 | 19.05 | 23.81 | 80.95 | 78.57 |
| **Bit-plane 3** | 8×8 | 09.52 | 14.29 | 14.29 | 14.29 | 88.10 | 85.71 |
|  | 16×16 | 14.29 | 19.05 | 14.29 | 23.81 | 85.71 | 78.57 |
|  | 32×32 | 14.29 | 19.05 | 14.29 | 14.29 | 85.71 | 80.95 |
| **Bit-plane 4** | 8×8 | 14.29 | 19.05 | 04.76 | 23.81 | 90.48 | 78.57 |
|  | 16×16 | 00.00 | 14.29 | 09.52 | 14.29 | 97.62 | 80.95 |
|  | 32×32 | 09.52 | 14.29 | 04.76 | 19.05 | 92.86 | 85.71 |
| **Bit-plane 5** | 8×8 | 14.29 | 19.05 | 04.76 | 14.29 | 90.48 | 85.71 |
|  | 16×16 | 19.05 | 09.52 | 19.05 | 19.05 | 80.95 | 85.71 |
|  | 32×32 | 09.52 | 19.05 | 09.52 | 14.29 | 90.48 | 80.95 |
| **Bit-plane** | 8×8 | 09.52 | 14.29 | 19.05 | 23.81 | 85.71 | 80.95 |
| 6 | 16×16 | 14.29 | 14.29 | 19.05 | 19.05 | 83.33 | 83.33 |
| | 32×32 | 14.29 | 19.05 | 23.81 | 14.29 | 80.95 | 85.71 |
| **Bit-plane 7** | 8×8 | 14.29 | 14.29 | 14.29 | 23.81 | 85.71 | 80.95 |
| | 16×16 | 19.05 | 19.05 | 14.29 | 14.29 | 83.33 | 83.33 |
| | 32×32 | 19.05 | 19.05 | 19.05 | 14.29 | 80.95 | 83.33 |

**Table 5** variation of performance rate FAR, FRR and accuracy with different block sizes for the second method

| | No of Block | Performance rate | FAR (%) | FRR (%) | Accuracy (%) | computational time in seconds |
|---|---|---|---|---|---|---|
| **Using 1st set of minutiae points** | 8×8 | 90.48 | 14.29 | 04.76 | 90.48 | 0.026918 |
| | 16×16 | 76.19 | 14.29 | 23.81 | 80.95 | 0.210892 |
| | 32×32 | 85.71 | 14.29 | 14.29 | 85.71 | 0.017936 |
| **Using 2nd set of minutiae points** | 8×8 | 95.24 | 00.00 | 14.29 | 92.86 | 0.136307 |
| | 16×16 | 83.33 | 19.05 | 14.29 | 83.33 | 0.021422 |
| | 32×32 | 78.57 | 14.29 | 23.81 | 80.95 | 0.017503 |

**Table 6** variation of performance rate FAR, FRR and accuracy with different block sizes for the third method

| | No of Block | Performance rate | FAR (%) | FRR (%) | Accuracy (%) | computational time in seconds |
|---|---|---|---|---|---|---|
| **Using 1st set of minutiae points** | 8×8 | 90.48 | 14.29 | 04.76 | 90.48 | 0.010010 |
| | 16×16 | 85.71 | 14.29 | 14.29 | 85.71 | 0.323892 |
| | 32×32 | 85.71 | 14.29 | 14.29 | 85.71 | 0.057366 |
| **Using 2nd set of minutiae points** | 8×8 | 85.71 | 14.29 | 14.29 | 85.71 | 0.145690 |
| | 16×16 | 85.71 | 14.29 | 14.29 | 85.71 | 0.126584 |
| | 32×32 | 85.71 | 14.29 | 14.29 | 85.71 | 0.658702 |